\documentclass{ieeeaccess}

\usepackage[hyphens]{url}
\usepackage[hidelinks]{hyperref}
\hypersetup{breaklinks=true}
\usepackage{cite}
\usepackage{amsmath,amssymb,amsfonts}
\usepackage{algorithmic}
\usepackage{graphicx}
\usepackage{textcomp}
\usepackage{afterpage}
\usepackage{array}
\usepackage{supertabular}
\usepackage{multirow}
\usepackage{multicol}
\usepackage{multirow}
\usepackage{afterpage}
\usepackage{siunitx}
\usepackage{color}

\def\BibTeX{{\rm B\kern-.05em{\sc i\kern-.025em b}\kern-.08em
    T\kern-.1667em\lower.7ex\hbox{E}\kern-.125emX}}
\begin{document}
\history{Date of publication xxxx 00, 0000, date of current version xxxx 00, 0000.}
\doi{10.1109/ACCESS.2022.DOI}

\title{Sample-Efficient Training of Robotic Guide Using Human Path Prediction Network}
\author{\uppercase{Hee-Seung Moon 
\uppercase{and Jiwon Seo}}, \IEEEmembership{Member, IEEE}}
\address{School of Integrated Technology, Yonsei University, Incheon 21983, Republic of Korea}
\tfootnote{
This work was supported in part by the Basic Science Research Program through the National Research Foundation of Korea (NRF) funded by the Ministry of Education (NRF-2018R1D1A1B07043580) and in part by the Unmanned Vehicles Core Technology Research and Development Program through the National Research Foundation of Korea (NRF) and the Unmanned Vehicle Advanced Research Center (UVARC) funded by the Ministry of Science and ICT, Republic of Korea (2020M3C1C1A01086407).
}

\markboth
{Moon \headeretal: Sample-Efficient Training of Robotic Guide Using Human Path Prediction Network}
{Moon \headeretal: Sample-Efficient Training of Robotic Guide Using Human Path Prediction Network}

\corresp{Corresponding author: Jiwon Seo (jiwon.seo@yonsei.ac.kr)}

\begin{abstract}
Training a robot that engages with people is challenging; it is expensive to directly involve people in the training process, which requires numerous data samples.
This paper presents an alternative approach for resolving this problem.
We propose a human path prediction network (HPPN) that generates a user's future trajectory based on sequential robot actions and human responses using a recurrent-neural-network structure.
Subsequently, an evolution-strategy-based robot training method using only the virtual human movements generated using the HPPN is presented.
It is demonstrated that our proposed method permits sample-efficient training of a robotic guide for visually impaired people.
By collecting only 1.5 K episodes from real users, we were able to train the HPPN and generate more than 100 K virtual episodes required for training the robot.
The trained robot precisely guided blindfolded participants along a target path.
Furthermore, using virtual episodes, we investigated a new reward design that prioritizes human comfort during the robot's guidance without incurring additional costs.
This sample-efficient training method is expected to be widely applicable to future robots that interact physically with humans.
\end{abstract}


\titlepgskip=-15pt

\maketitle

\section{Introduction}
\label{sec:introduction}

\IEEEPARstart{O}{ne} of the most important advantages that robots provide to human society is that they can complement the limited sensory and physical abilities of their users.
The use of a robotic guide, which offers navigational aid to visually impaired users, clearly illustrates this benefit.
Most visually impaired people have difficulty moving freely in unfamiliar spaces.
In the absence of visual recognition, users can rely on directional guidance from a robotic guide and achieve a safe route without being disrupted by unknown obstacles.
This type of assistive robot has various valuable applications, such as providing assistance to blind and elderly people with impaired visuomotor skills or guiding people from visually impoverished areas (e.g., during a disaster).

\begin{figure*}[t!]
\centering
\includegraphics[width=1.7\columnwidth]{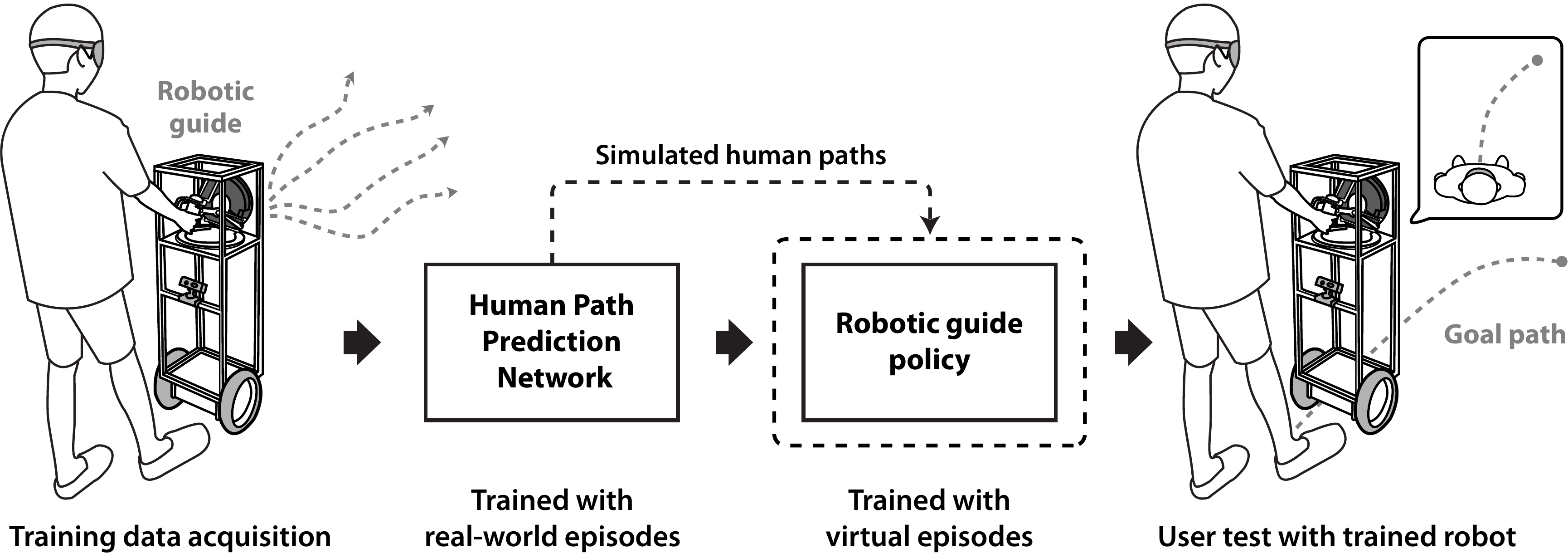}
\caption{Procedural overview of our training method for a robotic guide. We first train a human path prediction network with a limited number of episodes, and then train the robot policy using numerous virtual episodes generated based on the human path prediction network. The trained robotic guide is validated through a user test to guide users along the given goal path.
\label{fig:01}}
\end{figure*}

For robots engaging with humans (e.g., robotic guides), the ability to understand a user's behavior, that is, social intelligence, is essential~\cite{scheggi2016cooperative, sirithunge2019proactive, li2020data, moon2021optimal, antonucci2021efficient, moon2021fast}.
This allows robots to make highly sophisticated movements without reducing their usability and allows users to perceive more credibility from the robots.
A robotic guide with such social intelligence can accurately predict a user's movement and guide the user to follow a precise path.
This ensures the safety of blind people even in narrow and crowded places.
Furthermore, robot actions planned based on predicted human movements can lead to increased usability of a robotic guide, for example, inducing smooth user movements and avoiding excessive movements that put stress on the user's muscles.

In the past, social intelligence was usually provided to robots through handcrafted models~\cite{helbing1995social}.
However, recent deep learning-based computing methods have made it possible to provide robots with more advanced intelligence.
There are two representative methods for training a robot's action policy: reinforcement learning (RL) and evolution strategy (ES) methods.
On the one hand, RL methods develop the policy of an agent (e.g., a robot) by making the agent interact with the given task environment and gradually updating its policy (e.g., gradient descent).
On the other hand, ES methods find the optimal policy through a black-box optimization process that generates numerous candidate solutions and evaluates them.
By involving a neural network in an RL or ES method, recent studies have demonstrated significant results for training robots in terms of locomotion~\cite{peng2016terrain, hafner2020towards, bloesch2021towards}, object grasping~\cite{levine2018learning, lee2021beyond}, and finger dexterity~\cite{andrychowicz2020learning}.

However, there are critical problems in applying robot-training methods in real-world situations, particularly where humans are required to engage with robots.
The following two major problems are posed.
First, considerable amounts of time and spatial resources are required: While recent machine learning-based methods show promise, their sample efficiency is limited by the necessity of `trial and error' processes. 
Typical agent training using deep RL methods requires millions of data samples~\cite{levine2018learning, peng2016terrain}, making it practically impossible to collect such a large number of data samples when a human user is engaged.
Second, it is difficult to ensure the safety of the human partners participating in the training process.
The robot's movements during the early stage of the RL or ES training period are unrefined because the robot must go through an `exploration' process of taking various random actions and searching the sample space that can yield high cumulative rewards. 
Therefore, human involvement in such unpredictable robot movements can cause discomfort to users or, more seriously, lead to a dangerous situation.

In this paper, we present a training method for a robotic guide to compensate for both the above problems, as illustrated in Figure~\ref{fig:01}.
In particular, we propose a human path prediction network (HPPN) that predicts the next human movement based on the robot's sequential actions and the user's response data. 
The trained HPPN can generate an infinite number of virtual human movements. 
Therefore, a robotic guide can be trained using the ES method with numerous self-simulations based on virtual human movements. 
Our method does not directly involve people in training robot policies (RL or ES), but requires only a relatively small amount of movement data for training the HPPN. 
Moreover, our robot-training process is safe because the training dataset for the HPPN can be obtained from a general human guidance situation, and all inevitable trial-and-error processes for robot training are enacted only through a virtual simulation.

Another benefit of our approach is that we can repeatedly perform robot training according to different reward designs, without additional human involvement.
Using infinitely generated data based on an HPPN, various types of intelligence for a robotic guide pursuing different objectives can be acquired.
For example, one robotic guide policy might prioritize only the accuracy of the guidance (e.g., how precisely the assisted user follows the given path), while another robot policy might also take the comfortable movement of the assisted user into account in addition to the accuracy by shaping the training objective (i.e., reward).

To validate the proposed method, we developed a robotic guide testbed that physically guides users to follow a given target path without using their vision.
The users consistently receive kinetic guidance from the robot through a haptic device mounted on the robot.
Our robot observes the user's torso movements using a depth camera and measures the responsive kinetic force from the user using the haptic device. 
The HPPN was trained to predict the user's next movement based on multimodal response data, and then our robot learned its optimal behavior using the ES method from virtual simulations. 
We trained the robot using two different reward designs: one that pursues guidance performance (speed and accuracy) only, and another that pursues comfortable user movement as well as the guidance. 
In an indoor setting where a motion capture system can precisely measure user trajectories, we conducted a user test using our robotic guide. 
Our results demonstrated that, despite the limited amount of data collection, the trained robot could precisely guide the participants along the target path. 
Furthermore, among the two reward designs, the robot policy trained with a reward that additionally considered human comfort led to smoother movements of the actual participants.

\subsection{Contributions}
The contributions of this paper are summarized as follows:
\begin{itemize}
    \item We proposed a human path prediction network (HPPN) and training method for robots engaging with human partners based on the HPPN. Our training method compensates for two critical problems: sample inefficiency and human safety.
    \item We developed a robotic guide testbed for blind navigation that collects multimodal human-response data and physically guides users based on the sensed human data.
    \item We validated our trained robotic guide through a user test with eight participants. We demonstrate that our method is effective in developing human-centered robots, achieving smoother user motion and better goal-related metrics (e.g., speed and accuracy).
\end{itemize}

\section{Related Work}
\subsection{Robotic Aids for Blind Navigation}
Over the last few decades, the need for safe navigation for people with a visual impairment has been continuously raised. Since the pioneering development of robotic aids for blind navigation, such as NavBelt~\cite{shoval1998navbelt} and GuideCane~\cite{ulrich2001guidecane}, robotic assistive technologies have been steadily growing. NavBelt is a belt-typed robotic device equipped with ultrasonic sensors, which provides information regarding the detected obstacles around the user through acoustic feedback. Another type of robotic aid, GuideCane, is a mobile robot moving ahead of the user. Similar to the NavBelt, it detects obstacles through ultrasonic sensors and delivers the kinetic feedback to the user through an attached cane. Subsequent studies have utilized various advanced sensors, such as radio-frequency identification~\cite{kulyukin2006robot}, ultra-wideband systems~\cite{lu2021assistive}, laser scanners~\cite{scheggi2016cooperative, wachaja2017navigating, li2019toward}, and depth cameras~\cite{poggi2016wearable, wang2017enabling, kayukawa2019bbeep}; and their robotic guides provide common functionalities of obstacle detection and avoidance~\cite{kulyukin2006robot, bousbia2011navigation, poggi2016wearable, scheggi2016cooperative, wachaja2017navigating, wang2017enabling, li2019toward, lu2021assistive} and guidance along a given safe path~\cite{ghosh2014experience, chuang2018deep}.

Recent deep neural-network-based computing methods have been applied to the robotic guides, contributing especially to robot vision. Regarding obstacle detection functionality, Poggi and Mattocia~\cite{poggi2016wearable} developed a wearable device that detects and classifies objects ahead of blind users from depth images using the LeNet model architecture~\cite{lecun1998gradient} based on a convolutional neural network (CNN). Niu \textit{et al.}~\cite{niu2017wearable} presented another wearable device with the aim of assisting the blind users in opening a door. Using a stereo camera, the device detects the position of a doorknob and the hands of the users using a CNN-based model, and delivers audio feedback regarding which direction the users should move their hand. With regard to the function of guidance of a user along a given path, a robotic guide dog was developed by Chuang \textit{et al.}~\cite{chuang2018deep}, which recognizes a trail on the floor using a CNN-based architecture and determines the next robot motion along the trail.

Predicting human trajectory is another important factor to guide users in a precise and comfortable manner.
Especially, for the blind navigation scenario, it is challenging to predict human trajectories accurately because human movements are easily affected by several factors, such as the user's acceptance of a navigational aid~\cite{abdolrahmani2017embracing}, situational contexts~\cite{guerreiro2018context}, and environmental factors~\cite{kacorri2018environmental}.
There have been several machine learning-based approaches to analyze behavior of a user moving with a mobile robot and predict the user's next trajectory~\cite{yan2018data, moon2019prediction, li2020data, akabane2021pedestrian}.
However, how the learned user behavior model can be reflected in the navigational guidance of the robot and used to optimize the robot policy remains an open question.
For an example of non-learning-based methods, Scheggi \textit{et al.}~\cite{scheggi2016cooperative} developed a robotic guide that detects the user's position using a depth camera and reduces the moving speed to suit the user's intention. 
One limitation of~\cite{scheggi2016cooperative} is that the user's position data only applies to slowing and lifting the pace.
The robotic guide in our study uses a neural-network-based model to predict human movements and reflects this prediction data to select the next action of the robot for more precise and comfortable guidance.

\subsection{Training of Autonomous Robots}
With the development of artificial intelligence technology, there have been significant achievements in training robots to perform complex tasks on their own. RL algorithms have been perceived as a promising way to dealing with real-world robot manipulation tasks, such as, picking up objects~\cite{zhang2015towards, bohez2017sensor, lee2021beyond}, opening doors~\cite{gu2017deep, chebotar2019closing}, and locomotion~\cite{tan2018sim, bloesch2021towards}. In addition to RL methods, recent studies have suggested that ES methods can be an effective alternative for robot training~\cite{salimans2017evolution}. Although ES methods have an even lower sample efficiency than RL methods, ES methods are computationally efficient and have better exploration characteristics, which can provide robustness in robot training. Accordingly, a covariance matrix adaptation evolution strategy (CMA-ES)~\cite{hansen2006cma}, which is one of the latest ES methods, has been applied to robot tasks, such as the locomotion of humanoid~\cite{shafii2015learning} and quadruped~\cite{gehring2016practice} robots.

As a problem with the RL and ES methods, struggling to collect large amounts of data from real robots, researchers have attempted to acquire training data from the virtual robot movements on a physics-based simulator~\cite{tai2017virtual, tobin2017domain, tan2018sim, chebotar2019closing}. These simulator-based approaches have inherent errors between the simulation results and reality; however, researchers have succeeded in mitigating this problem by randomizing the simulation parameters~\cite{tobin2017domain} or adapting the parameters through a few real-world rollouts~\cite{chebotar2019closing}.

For the case of robots working with humans, the sample efficiency problem is still challenging because the collection of training data is more expensive and there is lack of physics-based simulators that fully embody the unpredictable nature of human behavior. Therefore, few studies applied RL or ES methods to train the robots that collaborate with people. Qureshi \textit{et al.}~\cite{qureshi2016robot} presented a humanoid robot that successfully learned when to extend its hand for a handshake with a human by applying a deep Q-network method. However, the robot had a long real-world training period of 14 days, showing the sample-inefficiency problem. Shafti \textit{et al.}~\cite{shafti2020real} also showed the possibility that the off-policy RL algorithm would be applied for robots to learn collaborative tasks with human partners in real-world environment.

A recent approach to addressing the sample efficiency problem is to model the movement of a user interacting with a robot, and then generate virtual data that can be used for robot training. Ghadirzadeh \textit{et al.}~\cite{ghadirzadeh2016sensorimotor} implemented a user's behavior model through Gaussian process and applied Q-learning approach to train a collaborative robot with modeled human behavior. Lathuili{\`e}re \textit{et al.}~\cite{lathuiliere2019neural} also pretrained a deep Q-network on generated human movement dataset to optimize the robot's gaze control during human--robot interaction, and then went through a fine-tuning process with real-world data. 

Besides, recent neural network-based generative models have been shown to be promising in synthesizing higher-dimensional data including realistic images. Using variational autoencoder (VAE)~\cite{kingma2013auto}, Ha and Schmidhuber~\cite{ha2018recurrent} succeeded in modeling complex environment surrounding an agent and training the agent only by simulated rollouts from the modeled environment. Their validation was limited to learning an agent that performs a video game in virtual environment; however, Thabet \textit{et al.}~\cite{thabet2019sample} later showed that the approach of Ha and Schmidhuber~\cite{ha2018recurrent} can be applied even in training real-world robots interacting with humans. Nevertheless, because the variation by Thabet \textit{et al.}~\cite{thabet2019sample} focused on learning the robot's movement based on a given single-step image input, it is still unclear whether the approach in~\cite{ha2018recurrent} would be effectively applicable to more realistic environment with time-series and multimodal human behavior data.

In this study, we present a sample-efficient learning method of the robotic guide that can provide safe navigation to real-world users, by optimizing the action policy of the robot using simulated rollouts.
To achieve this goal, we propose the HPPN to model the multimodal dynamics of human behavior when following a robotic guide, including walking trajectory, kinetic force of a hand, and body posture, in a time-series manner, and combine the HPPN with the approach in~\cite{ha2018recurrent}.

\section{Robotic Guide Testbed}
\begin{figure}[t]
\centering
\includegraphics[width=0.8\columnwidth]{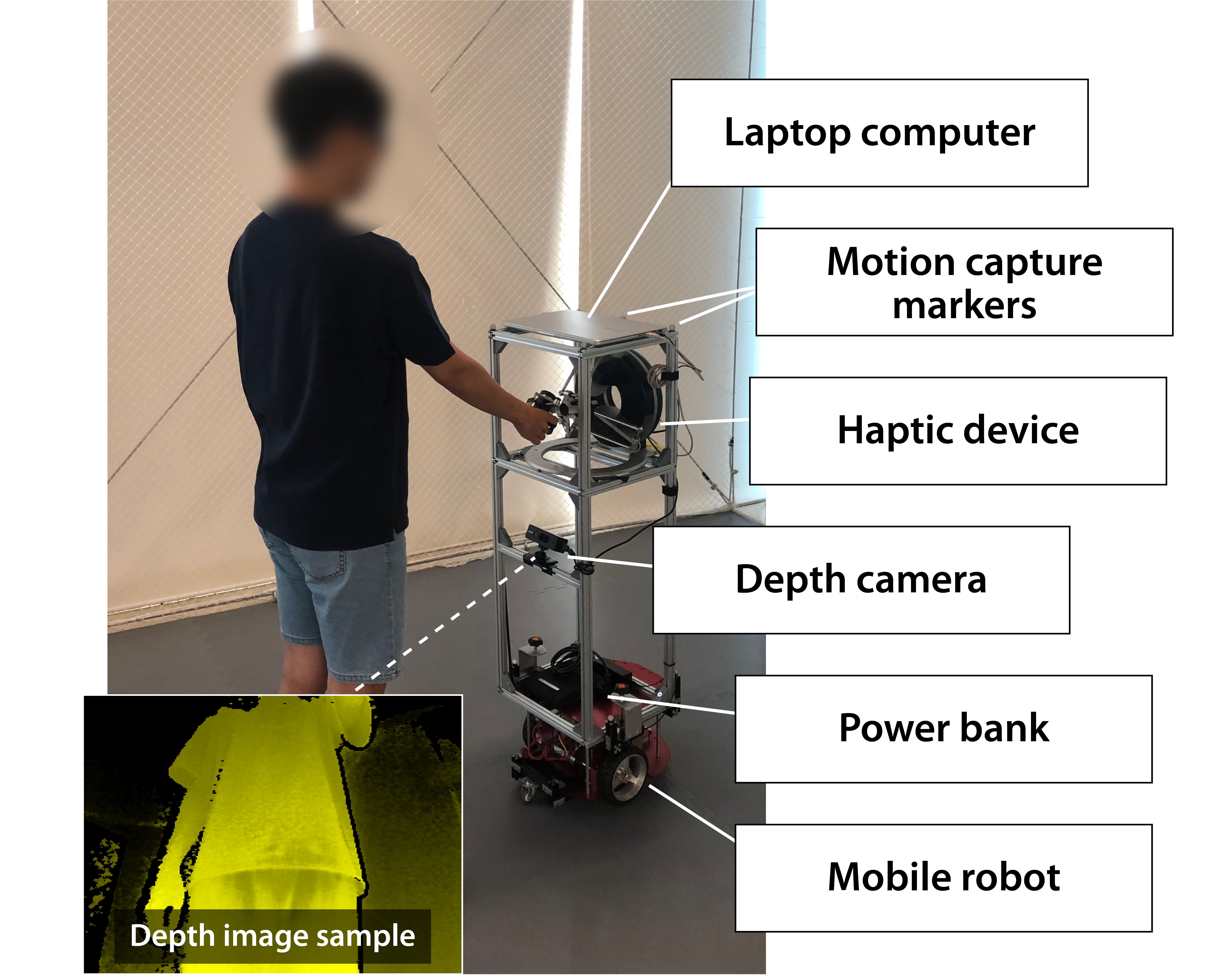}
\caption{Hardware overview of our robotic guide testbed.}~\label{fig:02}
\end{figure}

Our robotic guide is designed to guide the user to walk along a goal path by continuously delivering kinetic feedback to the user. As shown in Fig.~\ref{fig:02}, we mounted an Omega.7 haptic device (Force Dimension) on a Stella B2 mobile robot (NTREX, Inc.). To use our robotic guide, the user is instructed to grasp the handle of the haptic device, and the movements of the robot can be transmitted to the user directly through the haptic device. Another issue of our robotic guide design is the acquisition of multimodal human response data according to the robot's movements. As the first measurement data, a backward-facing Xtion 2 RGB-D camera (ASUS) is mounted on our robotic guide and captures the depth image of the user's torso at a pixel resolution of 640 $\times$ 480. For the second measurement data, we record the kinetic force input of the user from the haptic device. To quantitatively measure the force exerted on the robot, we apply a spring system that returns more strongly to the origin as the handle of the haptic device moves away from the origin. Specifically, we set the handle to move only in a 2D horizontal plane, and a 2D spring system toward the origin with a spring constant of 500 N/m is implemented in the haptic device. While using the robotic guide, the multimodal human response data are collected every 250 ms, i.e., four times a second. In addition, our robotic guide is equipped with a laptop computer, which controls the mobile robot and the sensors, motion capture markers for precise tracking of robot movements, and a power bank for powering the haptic device.

\section{Robot Training Method}
\subsection{Procedure}
Through our robot training procedure, the participants do not directly participate in the robot's RL- or ES-based learning process, which requires numerous data samples and cannot guarantee the safety of the users from the robot's trial-and-error process. Instead, the participants are instructed to follow along various safe movements of the robotic guide, and we train the HPPN using the collected dataset. Following the approach of Ha and Schmidhuber~\cite{ha2018recurrent}, we utilized VAE~\cite{kingma2013auto} structure to extract low-dimensional latent feature vectors from the high-dimensional depth images. Accordingly, we train the VAE in advance using the depth images in the training dataset, and the HPPN is then trained using robot actions, human kinetic force data, and compressed depth data as input values. The HPPN can act as a simulator providing the expected human path, given only the action sequence of the robotic guide. Using the HPPN-based simulator, we discover the optimal policy parameters for determining the next action of the robotic guide that maximizes our reward formulation using the CMA-ES algorithm. To summarize, our training method is applied in the following order:
\begin{enumerate}
\item Obtain a dataset of the users following the robot that guides them through various paths.
\item Train the VAE using the depth images in the dataset.
\item Train the HPPN using the dataset, including the latent feature vectors extracted using the trained VAE.
\item On the simulator based on the HPPN, optimize the policy parameters to maximize the cumulative reward of the robotic guide using the CMA-ES algorithm.
\end{enumerate}

\subsection{Data Acquisition}
\begin{figure}[t]
\centering
  \includegraphics[width=\columnwidth]{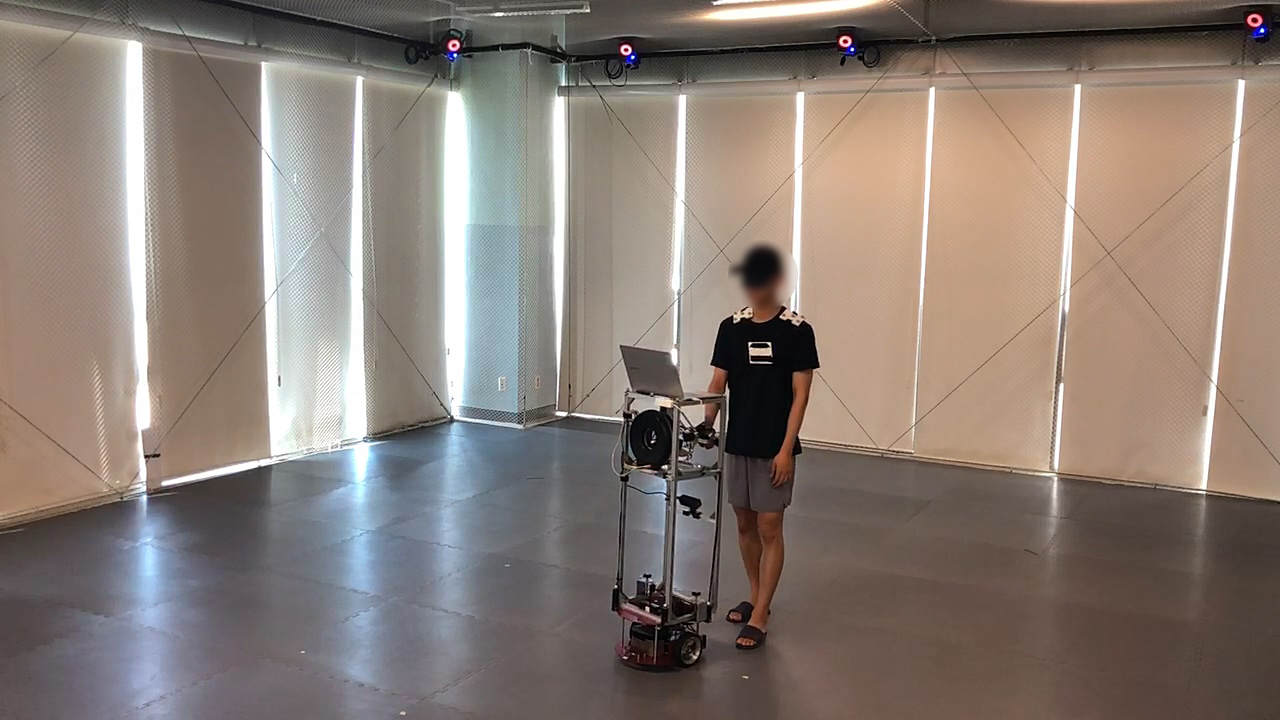}
  \caption{Experiment setup for training data acquisition. A blindfolded participant follows the robotic guide in an indoor environment with a motion capture system.}~\label{fig:03}
\end{figure}

We conducted a data acquisition session for training our HPPN. Since it is difficult to invite a sufficient number of blind participants, data were obtained from non-disabled participants with blindfolded, which is also common in previous robotic guide studies~\cite{scheggi2016cooperative, wachaja2017navigating, li2019toward}. In this session, 11 participants (2 females and 9 males) between the ages of 20 to 26 years (M = 23.64, SD = 1.67) were involved. All participants were right-handed. They were informed in advance regarding the purpose of the experiment and were free to rest when needed. During the session, blindfolded participants were instructed to move under the guidance of the robot in an open indoor environment without knowing the path of the robot in advance (Fig.~\ref{fig:03}). In the indoor environment, nine motion capture cameras were installed. Motion capture markers were attached on the robot and the participants; therefore, robot and human path data were collected with mm-level accuracy. Overall, the following time-series data were acquired: i) sets of human and robot movements, i.e., the change of the position and heading angle per timestep, ii) kinetic forces and depth images, measured from the robot sensors and iii) action commands given to the robot, i.e., goal speeds of the left and right wheels, for each timestep. 

According to Article 15 (2) of the Bioethics and Safety Act and Article 13 of the Enforcement Rule of Bioethics and Safety Act in Korea, a research project “which utilizes a measurement equipment with simple physical contact that does not cause any physical change in the subject” (Korean to English translation by the authors) is exempted from the approval. The entire experimental procedure was designed to use only a haptic device and a depth camera that did not cause any physical changes in the subject.

We aimed at obtaining human data based on as much diverse robot movements as possible, without drastic changes in the robot movements causing discomfort in the participants. To do so, the robot randomly chose its own left and right wheel speeds, but instead of changing the action for each timestep, it set the action to be maintained for arbitrary timesteps. In detail, the rotation speeds of the robot wheels were set between 2.5 to 5 rad/s (0.19 to 0.38 m/s), and the robot retained the action for 4 to 20 timesteps (for 1 to 5 s) before selecting the next action. This method allowed the robot to generate a smooth and diverse guidance route. One episode, which refers to one full guidance of the robot moving with a participant, took an arbitrary time of between 15 and 30 s. During the three-hour experiment, all participants were involved in 140 episodes. In total, 1,507 episodes of data were acquired, excluding data in which the robot and human paths were beyond the motion capture range and were therefore not fully measured.

\subsection{Human Path Prediction Network}

\begin{figure}[t!]
\centering
  \includegraphics[width=\columnwidth]{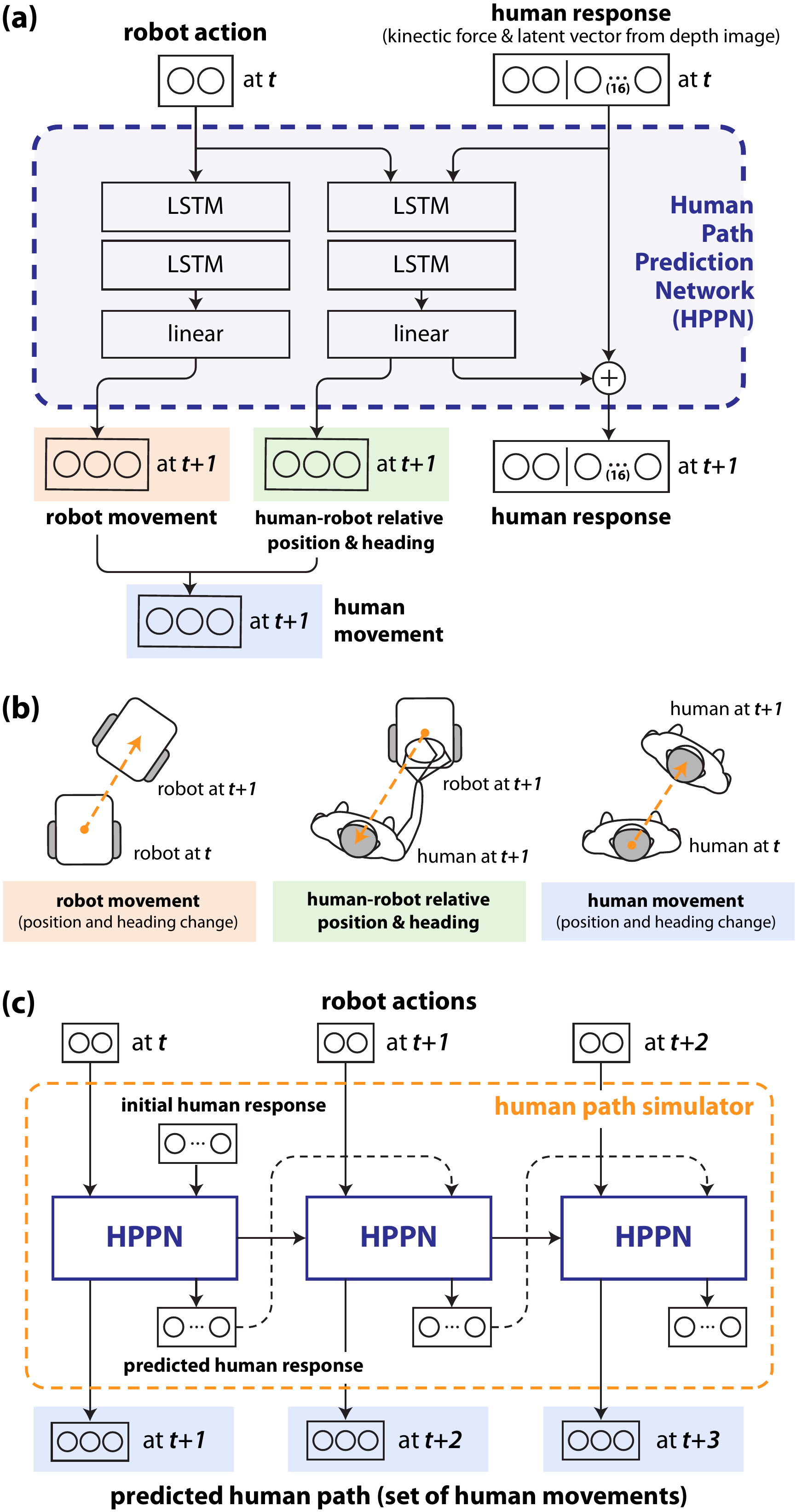}
  \caption{(a) Overview of our HPPN. The number of the circles in the box represents the size of each data. (b) Schematic representations robot movement, human--robot relative position and heading, and human movement. (c) The use of an HPPN as a simulator. The expected human path can be obtained by inputting the sequence of the robot actions.}~\label{fig:04}
\end{figure}

We implemented an HPPN for our robotic guide based on long short-term memory (LSTM) networks~\cite{hochreiter1997long}, an improved variant of recurrent neural networks. We extended the structure of the model described in our previous work~\cite{moon2019prediction}, which was used to predict only the relative position of a person from a robot. Fig.~\ref{fig:04} (a) shows an overview of our network. Our network employs the following input data: i) robot actions (two-dimensional), ii) kinetic forces applied to the haptic device (two-dimensional), and iii) latent feature vectors of depth images extracted from the pre-trained VAE (16-dimensional). We used the VAE with the structure described in our previous work~\cite{moon2019observation} to compress a depth image with a pixel resolution of 640 $\times$ 480 into a 16-dimensional vector.

\begin{figure}[t!]
\centering
  \includegraphics[width=1.0\columnwidth]{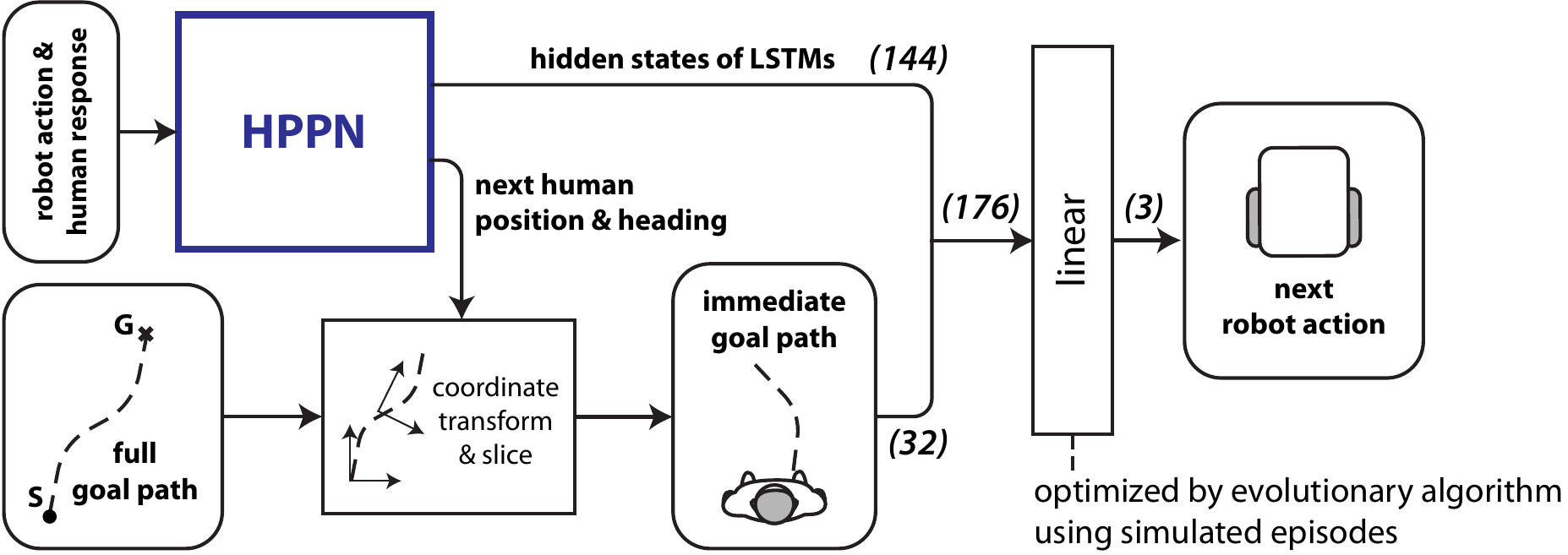}
  \caption{Process by which our robotic guide determines the next action according to the input data (current action, human response and goal path) and the policy parameters optimized using the human path simulator in Fig.~\ref{fig:04} (c). The numbers in parentheses indicate the dimensions of each data.}~\label{fig:05}
\end{figure}

Structurally, our network consists of two independent parts. One part is only responsible for the movement of the robot, and the other part is responsible for the movement of the user moving with the robot. The first part (left side of Fig.~\ref{fig:04} (a)) uses the robot's action only as an input, and outputs the next robot movement, i.e., the change of the position and heading per timestep, using two layers of the LSTM, which have eight hidden units each. Unlike the first part, which only deals with robot-related data, the second part (right side of Fig.~\ref{fig:04} (a)) focuses on predicting human motion using both the robot action and the human response, namely the kinetic force and latent vector from the depth image. Two layers of LSTMs with 64 hidden units are used to predict the human--robot relative position and heading at the next timestep. The network also predicts the next human response to create a simulator that predicts human movements using only the robot actions, as shown in Fig.~\ref{fig:04} (c), by applying a structure that uses the predicted human response as a new input for the next timestep. We applied a residual connection between the human response input and output, and therefore the network only predicts the amount of change, thus enabling more stable training. By obtaining robot movements in the first part and human--robot relative positions and headings in the second part, we acquire the predicted human movements as timestep progresses.

In detail, we used approximately 130K depth images in the training dataset to train the VAE, which was trained for 200 epochs using an Adam optimizer (learning rate of 0.001). For the HPPN, we trained the dual structures independently. By windowing the data with a size of 20 timesteps (i.e., 5 s), approximately 100K time sequential data samples were used to train the HPPN. Using the Adam optimizer at a learning rate of 0.01, the robot- and human-related LSTM structures were trained for 500 and 80 epochs, respectively.

\subsection{Training of Robotic Guide on Simulator}
We applied the CMA-ES method to train a robot policy with the generated episodes based on the human path simulator. The CMA-ES method is an evolutionary optimization algorithm that finds a solution that maximizes the objective function of a particular problem. During every generation, the method performs episodes with a population size of the candidate solutions and develops candidate solutions in a way that yields a better objective function over a generation. We implemented our robot policy using parameters that can be optimized using the CMA-ES method and defined rewards for the robot that can be used as the objective function, as described in the following sections. Using the human path simulator, a human path when the robot moves based on the current policy with the given parameters can be predicted. Under the CMA-ES method, the policy parameters improve using the reward calculated from the predicted human path, and this process is repeated to find the optimal policy parameters.

\subsubsection{Robot Policy}
During every timestep, our robotic guide determines the action to take in the next timestep by passing a robot's current feature vector through a single linear layer as follows:
\setlength{\arraycolsep}{0.0em}
\begin{eqnarray}
(next~robot~action~vector) = \mathbf{W} \cdot (feature~vector) + \mathbf{b},\nonumber
\end{eqnarray}
where $\mathbf{W}$ and $\mathbf{b}$ represent the weight and bias of the linear layer, and are the policy parameters optimized through the CMA-ES. The feature vector of the robot consists of two major features, as shown in Fig.~\ref{fig:05}. First, to contain the motion information of the user and the robot over time, we used the hidden state values of the LSTMs in the HPPN at the current timestep. Because the LSTMs were designed to predict the next human movement and affected by the sequential input data (robot actions and human responses), the hidden states of the LSTMs will integrate this information and are useful to reflect it in the robot's next action. Second, the information of the goal path that the user should walk along is provided to the robot to determine the next action. Instead of using a full goal path directly, only the immediate goal path for the current timestep is employed as the timestep passes. For this purpose, the coordinate transformation and slicing processes are performed at every timestep based on the predicted human position and heading obtained from the HPPN.

The feature vector we used has a size of 176, which is obtained by flattening the hidden state values of the LSTMs with a size of 144 (2 $\times$ 8 hidden units + 2 $\times$ 64 hidden units) and the immediate goal path data consisting of 16 two-dimensional points. The robot's action vector consists of three dimensions: the left motor speed, right motor speed, and degree ranging from 0 to 1 to determine whether to stop (the robot stops if the degree is over 0.5). Accordingly, our policy parameter has a size of 531, including a weight matrix with a size of 3 $\times$ 176 and a three-dimensional bias vector.

\subsubsection{Reward}
In this study, we set up two reward types: i) a reward that only considers goal efficiency (referred to as \textit{G Only}) and ii) a reward that also considers how the robot comfortably guides the user in addition to the goal efficiency (referred to as \textit{G + H}). In detail, the cumulative rewards of an episode based on these two reward types were calculated as follows:
\begin{itemize}
\item \textit{G Only}: This reward is determined based on the metrics related only to the goal efficiency, i.e., the completion time and accuracy. As a metric to measure the level of accuracy, the Fr\'echet distance, which calculates the similarity between two paths, was used to indicate how accurately the user actually moved along the goal path. The total calculation formula is as follows:
\begin{eqnarray}
(cumulative~reward) = -1\times(completion~time)\nonumber\\-100\times(Fr\acute{e}chet~distance).\nonumber
\end{eqnarray}
In addition, if the completion time exceeded the maximum timestep (which we set to 100 timesteps), the episode was terminated and a penalty of $-$500 was given.
\item \textit{G + H}: In addition to the metrics used for \textit{G Only}, we applied human motion smoothness to consider the comfort of the user. As a quantitative measure of the motion smoothness, we used the spectral arc length~\cite{balasubramanian2011robust}. The calculation formula is as follows:
\begin{eqnarray}
(cumulative~reward) = -1\times(completion~time)\nonumber\\-100\times(Fr\acute{e}chet~distance)\nonumber\\+30\times(spectral~arc~length~of~human~path).\nonumber
\end{eqnarray}
\end{itemize}

By setting up the two different rewards as above and evaluating them through user experiments, we can indicate the following two aspects. First, our trained HPPN have the advantage of training various types of robot policies that are optimal in different reward formulations by utilizing the fact that the simulator based on the HPPN can generate countless virtual human paths. Second, it can be confirmed whether the generated episodes based on the HPPN are effective for the development of real-world human-centered robots by comparing the actual performance of the two robot policies trained using the \textit{G + H} reward, which considers human convenience, and the \textit{G Only} reward, which does not.

\subsubsection{Implementation Details}
\begin{figure}[t]
\centering
  \includegraphics[width=\columnwidth]{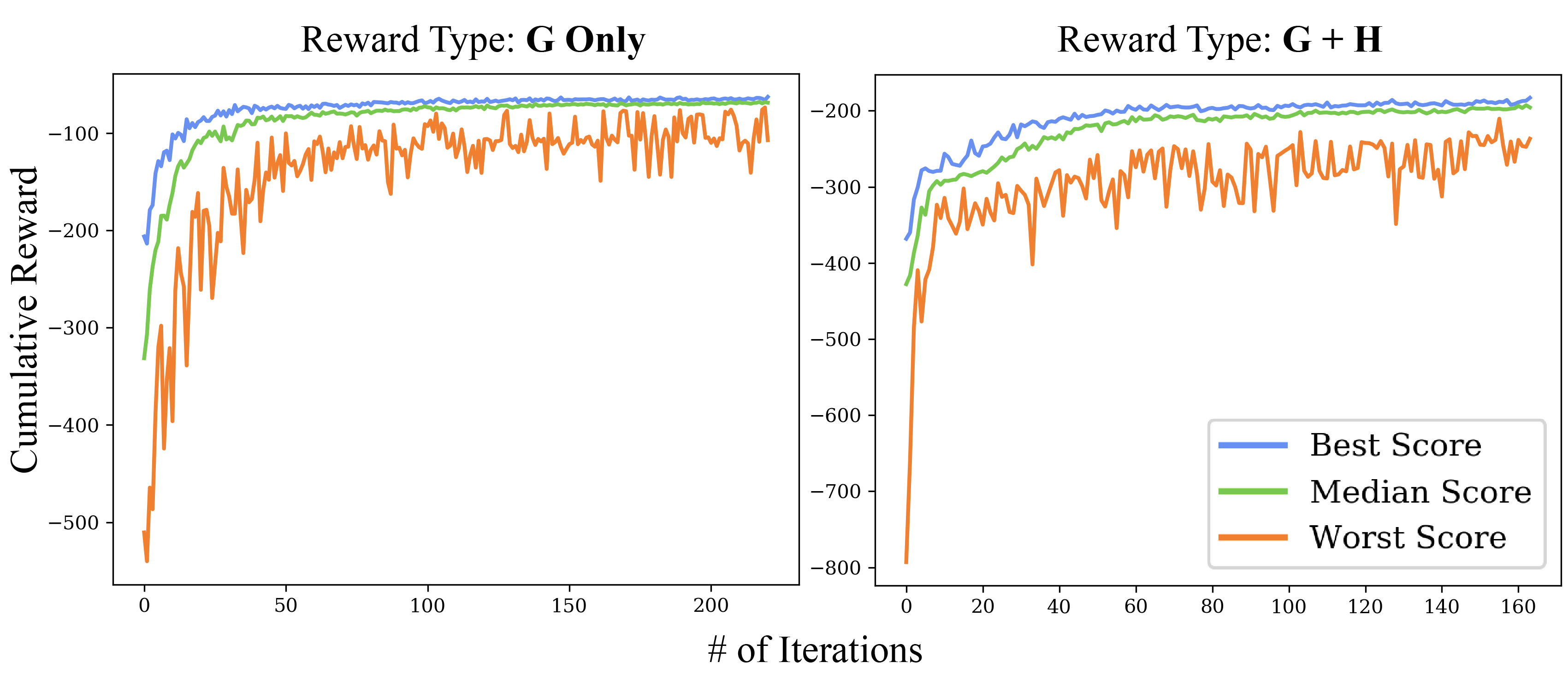}
  \caption{Convergence of the cumulative reward of our robot policy trained using the CMA-ES method with the \textit{G Only} (left) and \textit{G + H} (right) type rewards.}~\label{fig:06}
\end{figure}

For the CMA-ES method, we set 32 population sizes for each generation, which means that 32 different policy parameters were tested using the virtual simulations from one generation. For each policy parameter, we simulated 16 episodes based on 16 randomly generated goal paths of 4 m in length, and collected their average cumulative rewards. Fig.~\ref{fig:06} shows the robot policy training results for each reward type using this CMA-ES method. As a result of training applied until the objective functions of the best performer converged, it took 219 generations for \textit{G Only}, and 163 generations for \textit{G + H}. These figures indicate that 112,128 virtual episodes (219 $\times$ 32 $\times$ 16) were conducted for \textit{G Only}, and 83,456 (163 $\times$ 32 $\times$ 16) were conducted for \textit{G + H}. Considering that only 1,507 episodes were actually collected, our training method shows that the ES approach can be applied with a notably better sample efficiency.

\section{User Test}
\begin{figure}[t!]
\centering
  \includegraphics[width=0.8\columnwidth]{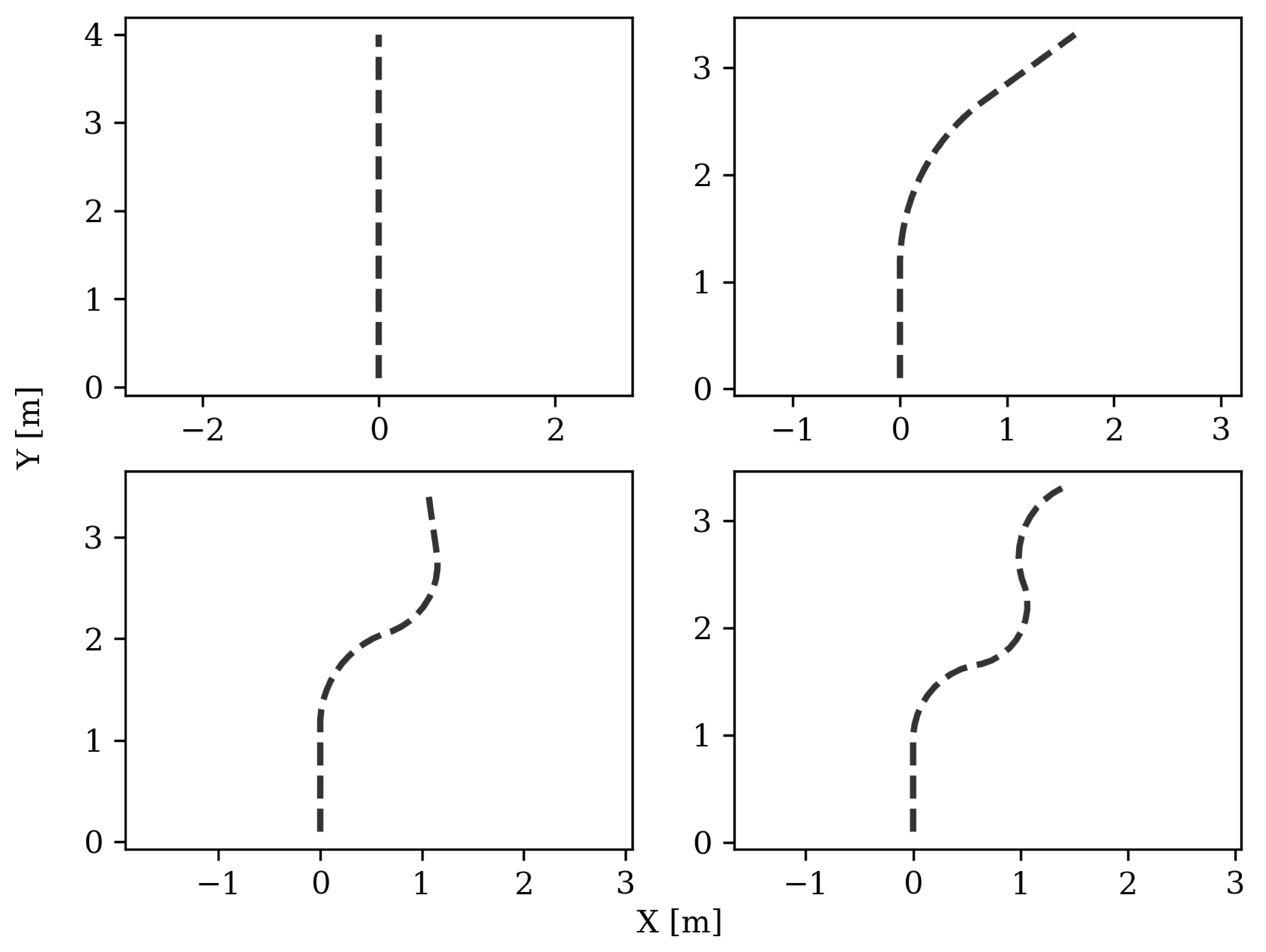}
  \caption{Goal paths used for the user test. All paths are 4 m long and created by adjusting the number of curvatures from zero to three. By laterally inverting the goal paths except for the straight line, seven goal paths were used for the user test.}~\label{fig:07}
\end{figure}

\begin{figure}[t]
\centering
  \includegraphics[width=\columnwidth]{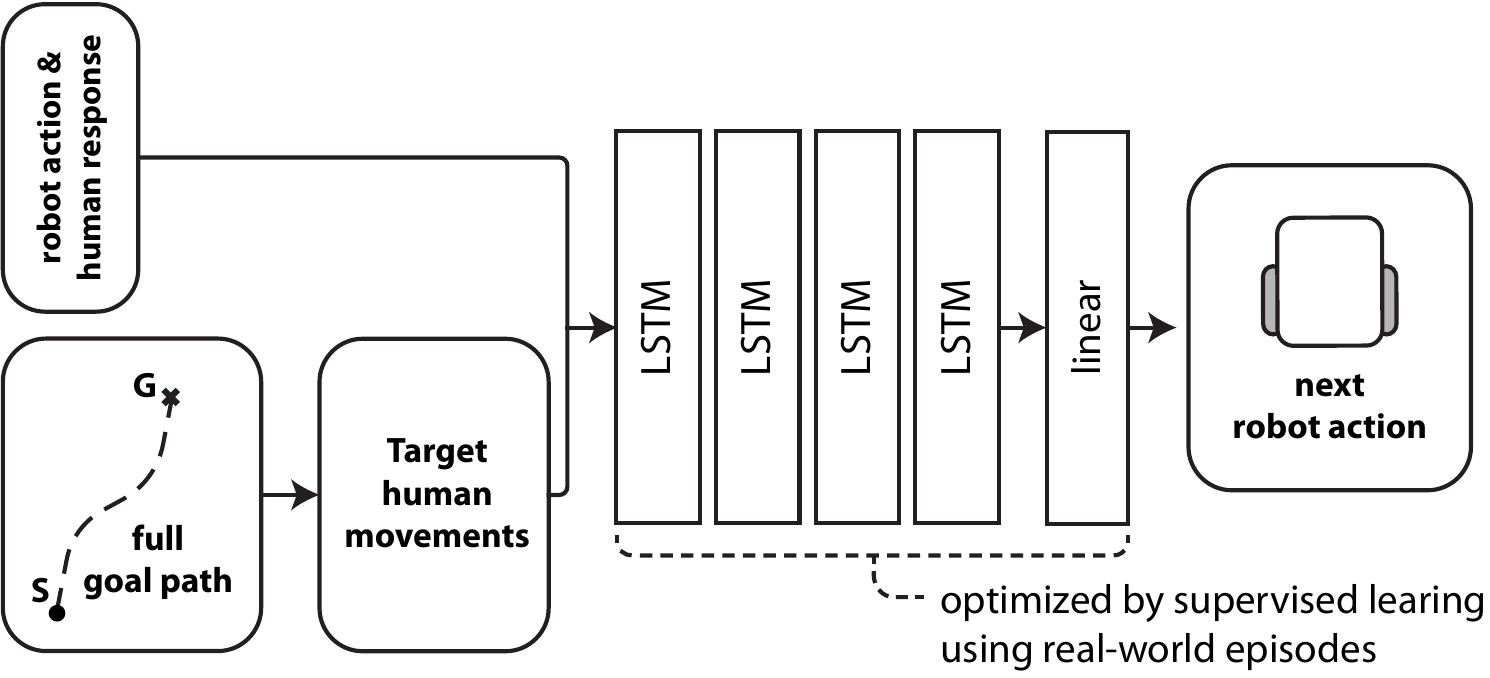}
  \caption{Overview of the baseline policy network.}~\label{fig:08}
\end{figure}

We conducted a user test to validate our robotic guide trained with the two reward types mentioned above. Eight participants (1 female and 7 males) between 24 to 29 years of age (M = 25.75, SD = 1.48) were involved in the user test. None of them participated in the training data acquisition session, and all were right-handed. Similar to the preceding training session, each participant was blindfolded and instructed to follow the robot's guidance without knowing the path information. Using the motion capture system, the precise path data of the participants were collected with mm-level accuracy to analyze the performance of our robotic guide. Note that the motion capture system that generated the ground truth was used only for the evaluation purpose during the user test. The trained robot does not utilize any information from the motion capture system when guiding the user. 

While the robot was set to take randomly generated movements during the training session, the trained robot policies were applied to guide the participants to follow the goal path during this user test. To investigate the performance of the robotic guide according to the path complexity, we generated various goal paths, controlling the number of curvatures in the path. As depicted in Fig.~\ref{fig:07}, we constructed 4 m long goal paths with zero to three curvatures. By laterally inverting the three goal paths except for the straight line, seven goal paths were used for the user test.

In this user test, three robot policies were evaluated: a baseline policy, described below, and our two robot policies trained using the \textit{G Only} and \textit{G + H} rewards. All participants conducted three episode trials for each of the 21 pairs of robot policy-goal path. Accordingly, each participant of this user test performed 63 episodes (7 goal paths $\times$ 3 robot policies $\times$ 3 trials) during a 1.5 h long experiment. Because the order of each robot policy-goal path pair was determined randomly, the participants were unable to know what conditions they were currently being guided under.

\begin{figure*}[t!]
\centering
  \includegraphics[width=1.8\columnwidth]{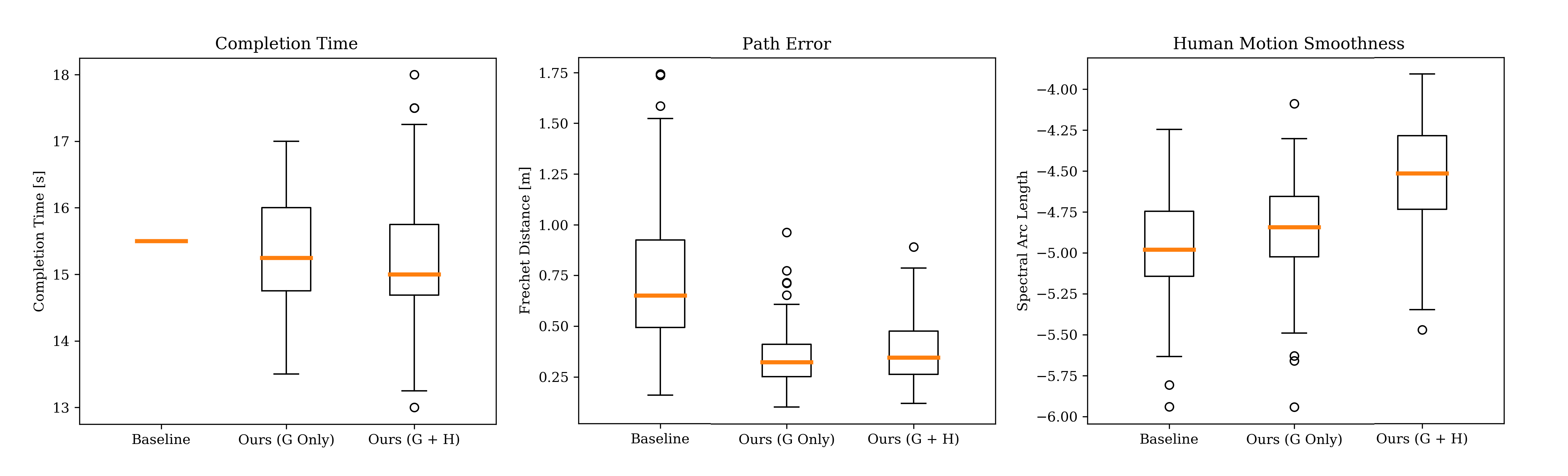}
  \caption{Box plot showing the overall performance results of each robot policy.}~\label{fig:09}
\end{figure*}

\subsection{Baseline Policy}
As a baseline policy, we developed a neural-network-based model that directly outputs the robot's next action from input data, which is the best way to train a robot with a limited amount of data and has shown effective performance in recent studies~\cite{chuang2018deep, nguyen2018translating}. As mentioned earlier, the conventional RL and ES methods suffer from the sample inefficiency problem. Thus, it is evident that the training with those methods will not be completed using the very limited number of real-world data samples used in this study. Therefore, a neural-network-based model, which can be trained using supervised learning even with the limited dataset we acquired, is implemented as our baseline policy. The performance of our training method based on the proposed HPPN was compared with that of the baseline policy.

The baseline policy network consists of four LSTMs with eight hidden units, as shown in Fig.~\ref{fig:08}. The network has the same input (current robot action, human response, and goal path) and output (next robot action) data as the robot policy we presented in Fig.~\ref{fig:05}. Because the training dataset consists of the actual distances that a person moved rather than a goal path, we delivered the goal path information in the form of target human movements that the robot should guide. Accordingly, an additional process that converts a given goal path into plausible target human movements is required for the user test. For the processing, we set the target speed of the user to 0.36 m/s, which corresponds to the top 80\% of the distribution of human speeds obtained from the training dataset.

\section{Results}
We evaluated the performance of the robotic guide from two perspectives: goal efficiency and user comfort. In terms of the goal efficiency, the completion time and path error were used as the evaluation metrics. The completion time of the robot guidance was measured as the time taken from the start to the stop of the robot. To quantify the path error, the Fr\'echet distance between the goal path and the actual path of the user was measured. For the user comfort, we focused on how comfortable the user could move under the guidance of the robot. A smooth movement of a person has been identified as the result of minimized effort~\cite{balasubramanian2015analysis, harris1998signal}. Therefore, we utilized the spectral arc length to measure the smoothness of the human movements.

\begin{figure}[t!]
\centering
  \includegraphics[width=0.9\columnwidth]{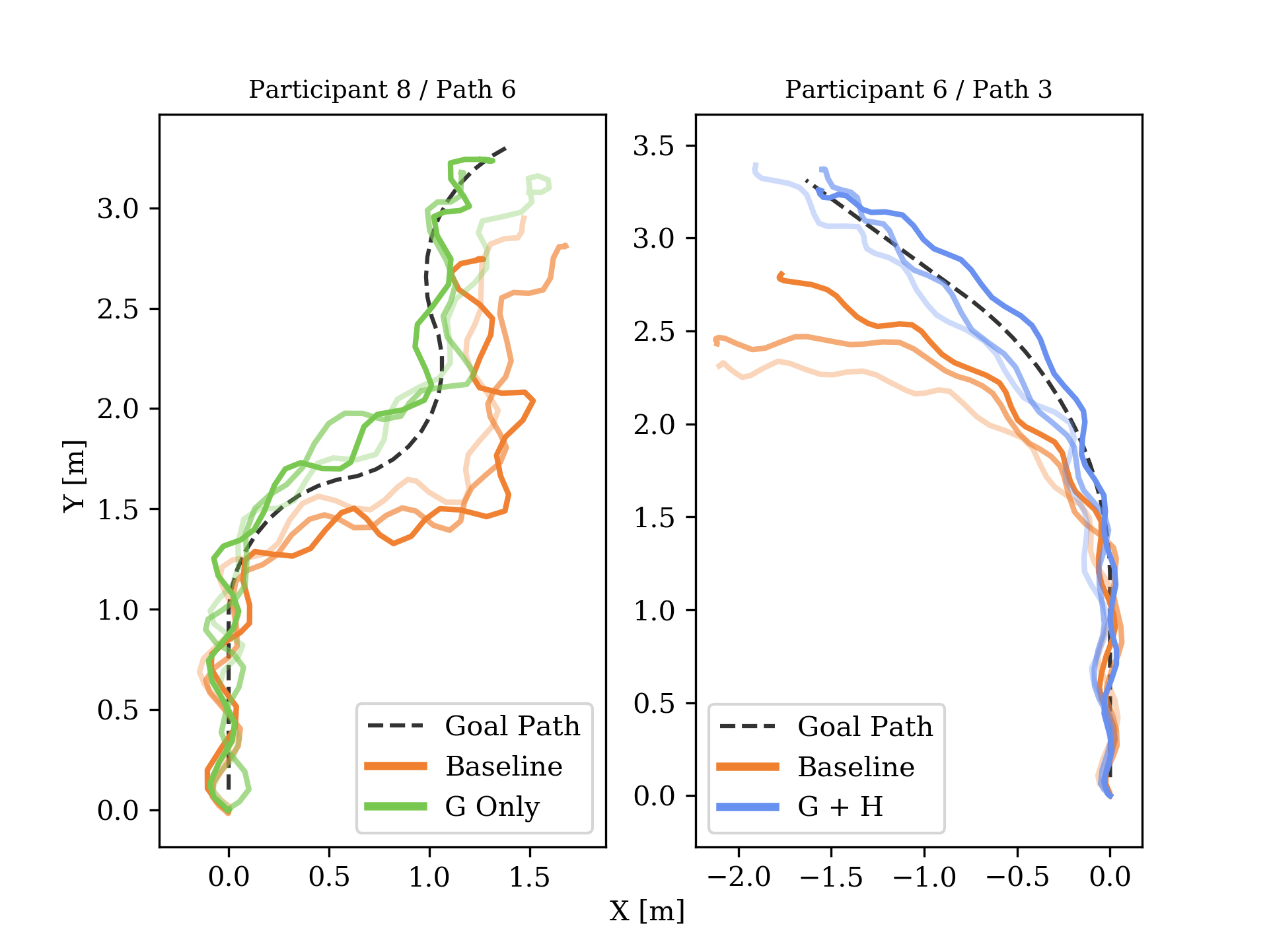}
  \caption{Samples of the user's actual path following the robot with the baseline policy and the \textit{G Only} (left) and \textit{G + H} (right) type policies. User paths during different trials are distinguished by the line transparency.}~\label{fig:10}
\end{figure}

Fig.~\ref{fig:09} shows box plots of the distributions of the above three metrics measured during all episodes of the user test. Through the paired t-test (eight participants), we verified whether one policy led to a significant difference in the guidance performance over another. In terms of the completion time, all three robot policies required a similar amount of time to guide the users. Note that, during the user test applying the same goal path length, the baseline policy consistently resulted in the same completion time. Because a fixed target user speed was utilized for the baseline policy, a fixed-length input data sequence was provided to the robot. Therefore, the robot with the baseline policy ended the guidance after the same amount of time. In terms of the path error, both the \textit{G Only} and \textit{G + H} type robot policies induced significant decrease of path error compared to the baseline policy  ($t = 9.931$, $p < 0.001$ and $t = 8.478$, $p < 0.001$, respectively). No significant differences were found between the \textit{G Only} and \textit{G + H} type robot policy ($t = 1.361$, $p = 0.216$). In terms of the smoothness of the human motion, the results of the \textit{G + H} type policy significantly outperformed those of both the \textit{G Only} type policy ($t = 8.236$, $p < 0.001$) and the baseline policy ($t = 12.975$, $p < 0.001$), whereas the \textit{G Only} type policy and the baseline policy did not show a significant difference ($t = 2.175$, $p = 0.066$). This indicates that the robot policy training method, which additionally considered motion smoothness, led to smoother movements of participants.

Fig.~\ref{fig:10} shows samples of the user's actual path when following the robot using the baseline policy and that of our robot policies (\textit{G Only} and \textit{G + H}). It is clearly shown that the user's path is closer to the dotted goal path, when using our robot policies. The oscillatory behavior of the user is a natural response to stepping on the left and right foot and has been reported in~\cite{winter1995human}. The trajectory of the robot does not have this oscillation, which is shown in Fig.~\ref{fig:11}.

\subsection{How the Robot Adequately Guides Different Users}

\begin{figure*}[t!]
\centering
  \includegraphics[width=2\columnwidth]{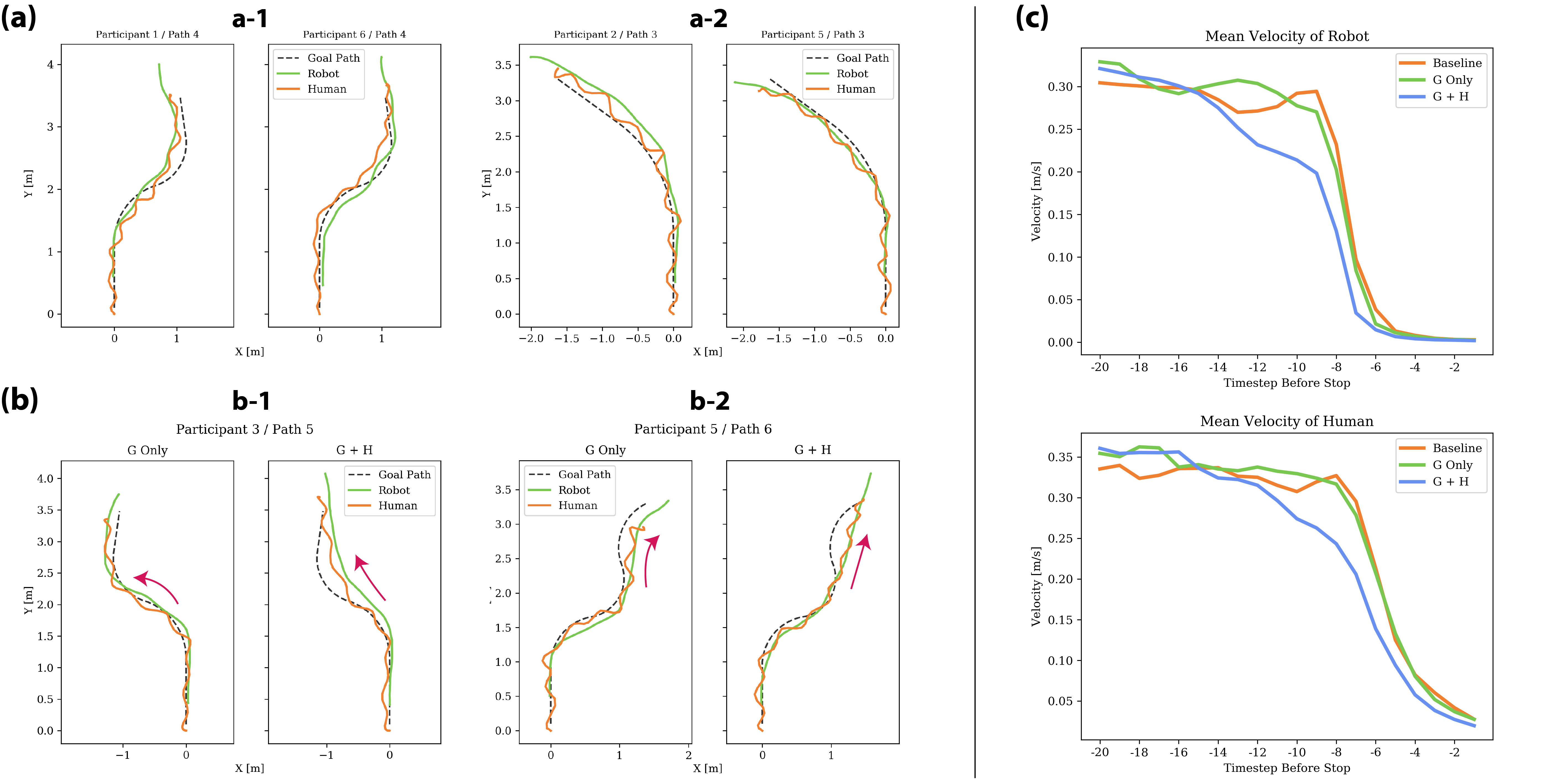}
  \caption{(a) Path samples of the robot using the \textit{G Only} policy when moving differently for different participants for the same goal path. (b) Path samples showing how robots under the \textit{G Only} and \textit{G + H} type policies behave differently for the same participant and the same goal path. (c) The mean velocity changes of the robot (top) and the user (bottom) before the robot stops.}~\label{fig:11}
\end{figure*}

Our robotic guide was designed to reflect human multimodal response data to better guide people to a goal path at each timestep. In other words, the robot learns to recognize the user's behavioral patterns from the multimodal response data and choose the best action based on the expected path of the user. We examined episodes from the user test to see how our robotic guide adequately guides different users.

Fig.~\ref{fig:11}(a) shows the path samples of the robots and humans during episodes in which the \textit{G Only} type robot policy guided two different participants to the same goal path. In the left sample (a-1), participants 1 and 6 showed different behavioral patterns in following the robot. Participant 1 tended to walk on the right side of the robot ahead, whereas participant 6 tended to walk on the left. Accordingly, our robotic guide showed different path movements for each participant. When guiding participant 1, the robot led to the left side of the goal path, and when guiding participant 6, the robot led to the right side of the goal path, consequently guiding the users closer to the goal path. These movements of the robot also appeared in another episode (a-2). While guiding along the same goal path, participant 2 walked along the left side of the robot, and participant 5 followed almost the same path as the robot. Accordingly, the different guidance paths of the robot were observed, moving toward the right side of the goal for participant 2, and moving as close as possible toward the goal path for participant 5. These episodes show that our trained robot effectively learned the different behavioral patterns of humans and chose adequate actions for different users.



\subsection{How the Robot Comfortably Guides Users}
During this user test, the \textit{G + H} type robot policy showed a superior smoothness of the human motion compared to the baseline and \textit{G Only} type policies. We further analyzed how the \textit{G + H} type policy, which was trained on virtual simulations, improved the motion smoothness of the actual users. Our analysis proceeded in two ways, in terms of the actual guiding path and the change in speed of the robot.

First, we compared the actual robot paths generated by the \textit{G Only} and \textit{G + H} type robot policies for the same participants based on the two path samples in Fig.~\ref{fig:11}(b). In both path samples, we observed that the \textit{G + H} type robot gave up turning sharply and chose a less convoluted path, although a sharp turn was required to follow the sharply turning goal path. In the path samples on the left (b-1), the \textit{G + H} type robot rotated only slightly along the goal path, whereas the \textit{G Only} type robot rotated obviously toward the left along the goal path (as depicted by the red arrows). In the path samples on the right (b-2), the \textit{G + H} type robot gave up the last turn and went straight, whereas the \textit{G Only} type robot moved along the goal path to the end (as depicted by the red arrows). These examples demonstrate that the \textit{G + H} type policy learned to avoid sharp rotations that can lower the smoothness of the human motion, even if a certain level of accuracy is lost. However, such movements of the \textit{G + H} type robot that noticeably reduce accuracy as in Fig.~\ref{fig:11}(b) were rarely observed and, as shown in Fig.~\ref{fig:09}, did not significantly reduce guidance accuracy compared to the \textit{G Only} type robot in general.

For the second aspect, we analyzed for all three policies the mean velocities of the robot and the user for 20 timesteps (i.e., 5 s) before the robot stopped, as shown in Fig.~\ref{fig:11}(c). The analysis showed that there were no noticeable differences between the baseline and \textit{G Only} type policy; however, the \textit{G + H} type robot started decelerating earlier than the other two policies, and gradually slowed down. The other two policies started the deceleration of the robot at 8 to 10 timesteps prior to stopping, whereas the \textit{G + H} type policy started the deceleration at 12 to 14 timesteps before it stopped. Accordingly, the speed of the user following the robot also decreased early and slowly under the \textit{G + H} type policy as compared to the other two policies. Consequently, this change in speed demonstrates that the \textit{G + H} type policy increased the smoothness of the human motion by learning its own gradual slowdown process prior to stopping.

\section{Conclusion}
In this paper, we introduced the HPPN, a neural network model that predicts human movements with regard to sensed human response data, and the training method for the robotic guide based on episodes generated by the HPPN.
Our approach is beneficial because it addresses concerns regarding human safety and sample inefficiency that arise when training robots to collaborate with humans.
We collected 1,507 real-world episodes for training the HPPN, and then it was possible to generate over 100,000 virtual episodes to optimize the action policy of the robotic guide.
The user test results indicated that our method is effective in training the robotic guide with increased guidance accuracy compared with the baseline method that used the same amount of real-world training data. 
In addition, using infinitely generated episodes, we can investigate various reward formulations to achieve a highly human-centered robot policy. 
For example, one of the remarkable points of our work is that, by utilizing the reward formulation that values human comfort, the trained robotic guide actually yielded improved smoothness in human motion during a real-world user test.

This study has several promising future extensions.
First, it is possible to improve the guidance performance of the robotic guide by utilizing additional sensors.
For example, the robotic guide in this study estimated its own location based only on its past action commands.
If the robot is combined with improved localization systems, more accurate guidance can be achieved over longer distances than those in the present study.
Second, future studies can investigate methods for providing personalized guidance according to different individual user characteristics. 
If the HPPN can predict an individual user's movement (e.g., the elderly or children) rather than general user movement by model adaptation (e.g., \cite{moon2021optimal, moon2021fast}), it is possible to optimize the robot's policy personalized for the user. 
Third, it is worthwhile to study various reward formulations that can improve the development of human-centered robots. 
Our focus in this study was on the smoothness of human movement, but there can be a variety of different metrics indicating human comfort (e.g., the amount of kinetic force applied to a user). 
Additionally, it is possible to train a generalized robot policy over various reward formulations using recent multi-objective RL~\cite{abels2019dynamic, yang2019generalized}. 
With the generalized robot policy, it is easy to investigate the reward formulation that users prefer, as there is no need to re-train the policy from scratch for each reward formulation. 
Finally, our sample-efficient robot-training method can be applied to other collaborative robots. 
It would be of great value to examine the viability of our approach in a variety of collaborative situations where robots need to interact with humans, such as in the case of industrial or surgical-assistive robots.

Recent growth in computing technology has made intelligent systems universal in our daily lives; it has become commonplace for robots or conversational agents to interact with people. 
In this regard, our research can contribute to the development of socially intelligent systems that can comprehend human behavior and discern user intentions.


\bibliographystyle{IEEEtran}
\bibliography{mybibfile, IUS_publications}

\phantomsection

\begin{IEEEbiography}[{\includegraphics[width=1in,height=1.25in,clip,keepaspectratio]{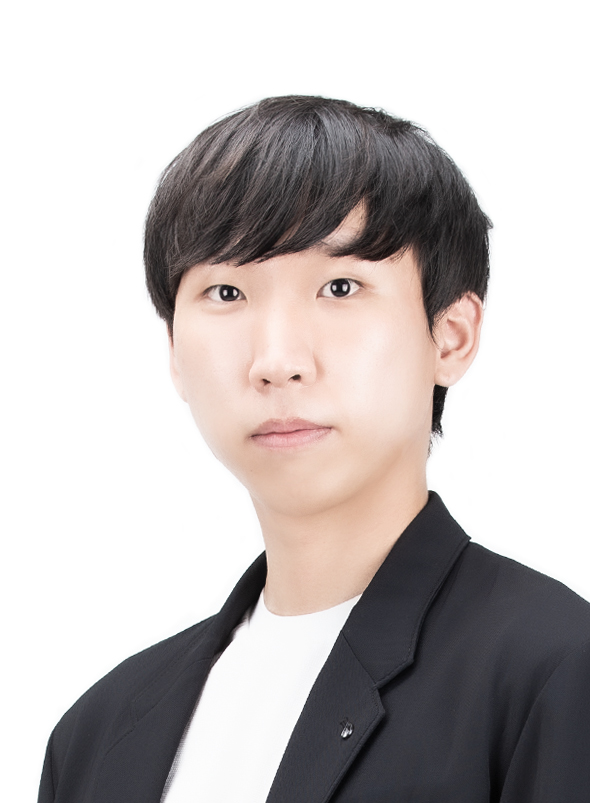}}]{Hee-Seung Moon} is a Ph.D. candidate in the School of Integrated Technology, Yonsei University, Incheon, Korea. He received the B.S. degree in Integrated Technology from Yonsei University in 2015. His research interests include human--computer interaction, computational interaction, user behavior modeling, and deep learning.
He received the Undergraduate and Graduate Fellowships from the Information and Communications Technology (ICT) Consilience Creative Program supported by the Ministry of Science and ICT, Korea.
\end{IEEEbiography}

\begin{IEEEbiography}[{\includegraphics[width=1in,height=1.25in,clip,keepaspectratio]{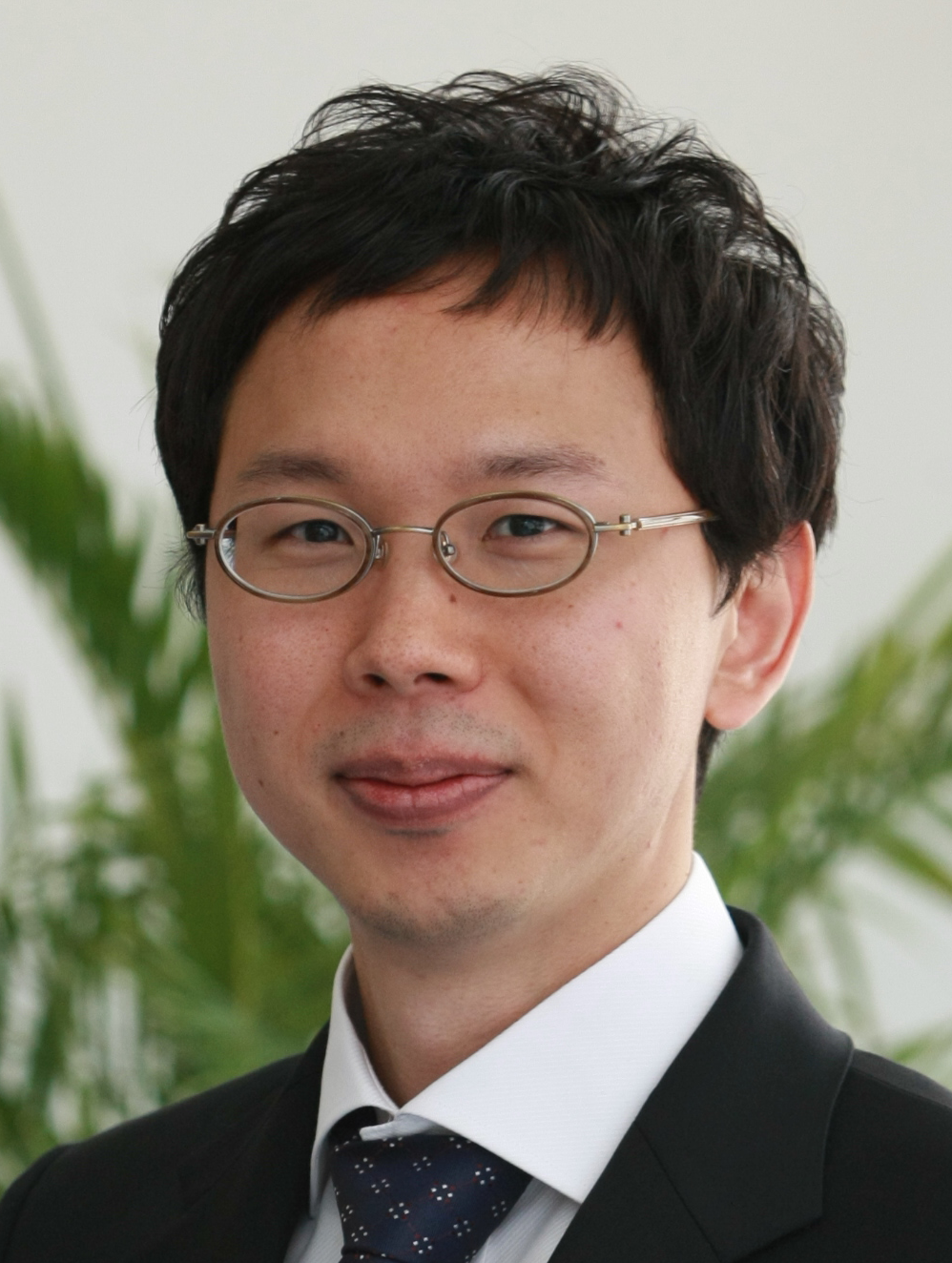}}]{Jiwon Seo} (M'13) received the B.S. degree in mechanical engineering (division of aerospace engineering) in 2002 from Korea Advanced Institute of Science and Technology, Daejeon, Korea, and the M.S. degree in aeronautics and astronautics in 2004, the M.S. degree in electrical engineering in 2008, and the Ph.D. degree in aeronautics and astronautics in 2010 from Stanford University, Stanford, CA, USA. He is currently an associate professor with the School of Integrated Technology, Yonsei University, Incheon, Korea. His research interests include GNSS anti-jamming technologies, complementary PNT systems, and intelligent unmanned systems. Prof. Seo is a member of the International Advisory Council of the Resilient Navigation and Timing Foundation, Alexandria, VA, USA, and a member of several advisory committees of the Ministry of Oceans and Fisheries and the Ministry of Land, Infrastructure and Transport, Korea.
\end{IEEEbiography}

\EOD

\vfill

\end{document}